\documentclass[letterpaper]{article} %
\usepackage{aaai2027}  %
\usepackage[hyphens]{url}  %
\usepackage{graphicx} %
\usepackage{natbib}  %
\usepackage{caption} %
\usepackage{algorithm}
\usepackage{algorithmic}
\usepackage{amsmath}
\usepackage{multirow}
\usepackage{array}
\usepackage[most]{tcolorbox}

\newtcolorbox{promptbox}[1]{colback=white,colframe=black!60,coltitle=black,
  colbacktitle=black!8,fonttitle=\bfseries\small,title={#1},
  boxrule=0.5pt,arc=1pt,left=4pt,right=4pt,top=3pt,bottom=3pt,fontupper=\small}

\usepackage{newfloat}
\usepackage{listings}
\DeclareCaptionStyle{ruled}{labelfont=normalfont,labelsep=colon,strut=off} %
\floatstyle{ruled}
\newfloat{listing}{tb}{lst}{}
\floatname{listing}{Listing}

\usepackage{booktabs}

\title{BAIT: Can LLMs Find What Biases a Reader?\\ A Benchmark for Cognitive Bias Triggers in News}
\author{
    Written by AAAI Press Staff\textsuperscript{\rm 1}\thanks{With help from the AAAI Publications Committee.}\\
    AAAI Style Contributions by Peter Patel Schneider,
    Sunil Issar,\\
    J. Scott Penberthy,
    George Ferguson,
    Hans Guesgen,
    Francisco Cruz\equalcontrib\corresponding,
    Marc Pujol-Gonzalez\equalcontrib\corresponding
}
\affiliations{
    \textsuperscript{\rm 1}Association for the Advancement of Artificial Intelligence\\

    1101 Pennsylvania Ave, NW Suite 300\\
    Washington, DC 20004 USA\\
    proceedings-questions@aaai.org
}

\title{BAIT: Can LLMs Find What Biases a Reader?\\ A Benchmark for Cognitive Bias Triggers in News}
\author {
    Yousra Fettach \textsuperscript{\rm 1}\corresponding,
    Tijl De Bie\textsuperscript{\rm 1}
}
\affiliations {
    \textsuperscript{\rm 1}Ghent University, Belgium\\
    yousra.fettach@ugent.be
}

\begin{document}
\nocopyright  

\maketitle

\begin{abstract}
Persuasive writing exploits cognitive biases, recruiting a reader's mental shortcuts to shape belief, often without the reader noticing. We call the linguistic features exploiting such mental shortcuts \emph{cognitive bias triggers}. As large language models (LLMs) increasingly mediate how people encounter news, summarizing it, ranking it, and answering questions over it, a natural question is whether they can recognize the features of a text that bias a human reader. We introduce BAIT (\textbf{B}ias \textbf{A}nnotation \textbf{I}n \textbf{T}ext), a resource for detecting cognitive bias triggers in news. BAIT pairs a schema of nine triggers, each specified in an annotator codebook, with 50 English news articles annotated by trained annotators at the article and span levels and adjudicated into a gold standard. On this data we define two tasks, detecting which triggers a text contains and locating the spans that carry them. We then benchmark six frontier LLMs zero-shot and with three or ten in-context examples. The models approach human agreement on which triggers an article contains, but rarely recover its full set of triggers, and they reach less than half of human agreement in locating the passages that carry them. Moreover, they over-predict triggers that come with easy surface cues, and most of the passages they mark for these triggers are not triggers, yet carry the trigger's surface cue far more often than random text. These findings suggest that the evaluated LLMs respond to surface cues rather than to the mechanisms that make a passage manipulative. We release the schema, the codebook, standoff annotations, and the evaluation pipeline.
\end{abstract}

\section{Introduction}
Cognitive biases are systematic deviations from rational reasoning that arise from the heuristic processes underlying everyday cognition \cite{tversky1974judgment}. These heuristics help readers process information efficiently. However, they can also be exploited through persuasive writing.

Such writing recruits these cognitive heuristics to shape belief and form attitudes \cite{petty1986elaboration}. Building on work on media bias and persuasion techniques in computational social science \cite{spinde2021neural, piskorski2023semeval}, we refer to such persuasive writing features as \emph{cognitive bias triggers}.

Cognitive bias triggers are hard to annotate because they cannot be identified from wording alone. Quoting an expert, recounting one person's story, or restating a claim are common in news writing, and each can appear in a passage that is a trigger and in one that is not. Telling the two apart depends on the rest of the article, for example whether the expert's claim is supported by evidence anywhere in the text. Work on the creation of annotated datasets for closely related phenomena, such as propaganda \cite{da2019fine} and persuasion techniques \cite{piskorski2023semeval}, reports the same difficulty. Both of these datasets relied on professional or expert annotators, detailed guidelines, and a consolidation step in which annotators resolved their disagreements. These datasets, however, do not record which cognitive bias a passage engages in the reader, or whether the conditions for that bias are met. To the best of our knowledge, no existing resource is annotated for this, so it is not yet possible to test whether a system can identify the passages that would bias a reader.

This question has become pressing as large language models (LLMs) increasingly stand between readers and the news, summarizing, retrieving, ranking, and answering questions about it. A model in this role decides what readers see and how it is framed. If it cannot recognize the passages that would bias a reader, it may pass them on unexamined or reproduce them in its own output.

A growing literature establishes that LLMs are themselves subject to human-like biases in their own judgments, including anchoring, framing, and order effects \cite{echterhoff2024cognitive,itzhak2024instructed,koo2024benchmarking}. That work characterizes the model as a biased reasoner. It leaves open whether a model can recognize, in text supplied to it, the features that would bias a human reader, and localize them precisely enough for the judgment to be inspected.

\paragraph{Statement of Contributions}
We introduce BAIT (\textbf{B}ias \textbf{A}nnotation \textbf{I}n \textbf{T}ext), a resource and evaluation pipeline for detecting cognitive bias triggers in news text. Our contributions are as follows.

\begin{enumerate}
\item We define a schema of nine cognitive bias triggers, each specified in an annotator codebook. Two trained annotators used it to annotate 50 English news articles at the article and span levels, and the annotations were consolidated into a gold standard. We release the schema, codebook, standoff annotations, and evaluation pipeline.\footnote{\url{https://github.com/aida-ugent/BAIT}}
\item We define two tasks, detecting which triggers an article contains and locating the spans that carry them, and benchmark six frontier LLMs zero-shot and with three and ten in-context examples.
\item We show that models can often tell that an article contains triggers but cannot reliably point to the passages that carry them: the best model reaches a trigger-level $F_1$ of 0.22, against 0.50 between two annotators. In-context examples narrow this gap without closing it.
\item We find that models respond to the surface cue of a bias rather than to its mechanism. They over-predict the triggers with the easiest cues, namely Negativity Bias (39 articles on average against 23 in the gold standard), Authority Bias (30 against 19), and Declinism (10.5 against 6). Most of the passages they mark for these triggers are not triggers, yet carry the trigger's surface cue far more often than random text.
\end{enumerate}
No part of this paper has been published before.

\section{Related Work}
\paragraph{Media bias and lexical bias detection.}
A substantial line of work annotates bias in news and locates it in word choice. BABE \cite{spinde2021neural} provides expert annotations of 3{,}700 news sentences, marking whether each sentence is biased and which words carry the bias, with higher inter-annotator agreement than earlier crowdsourced datasets such as MBIC \cite{spinde2021mbic}. MBIB \cite{wessel2023introducing} assembles nine media-bias tasks over twenty-two datasets, spanning linguistic, political, framing, hate-speech, and other bias types, to test how detection methods generalize across them. MBIB also includes a cognitive-bias task, which it defines as readers' selective choice of which sources to read and trust. In practice, however, the labels come from proxies: the political leaning of an article's outlet using the BIGNEWS dataset \cite{liu2022politics} and the fact-checked truthfulness of a statement with the LIAR dataset \cite{wang2017liar}.
\paragraph{Persuasion and propaganda techniques.}
A second line annotates the rhetorical devices a text uses to persuade. \citet{da2019fine} annotate news articles at fragment level with eighteen propaganda techniques and cast the problem as joint detection and localization. The SemEval-2023 Task~3 shared task \cite{piskorski2023semeval} extends this to twenty-three persuasion techniques across nine languages, adding genre and framing labels. Both datasets label the rhetorical devices a writer uses to persuade. Neither says which cognitive bias a passage might trigger in a reader, and a cue that relies on no named technique falls outside their label sets.
\paragraph{Cognitive bias in LLMs.}
A growing body of work asks whether language models display human-like biases in their own outputs: anchoring and framing effects in decision tasks \cite{echterhoff2024cognitive}, biases that emerge from instruction tuning \cite{itzhak2024instructed}, and order and egocentric effects when models act as evaluators \cite{koo2024benchmarking}. This work treats the model as a reasoner whose judgments can be distorted, and measures that distortion. A separate strand annotates reasoning defects in text directly, closest among them logical fallacy detection \cite{jin2022logical}, which labels invalid argument structure as a property of the argument itself.

Across these lines of work, the question of how a text might bias its reader remains open. No resource marks the passages likely to trigger a cognitive bias, or which bias each one triggers.

\section{The BAIT Schema}
\label{sec:schema}

BAIT targets cognitive biases that news text can \emph{trigger} in a reader:
well-documented tendencies in human judgment that a passage can set off through
the way it presents information. We selected candidate biases from the
psychological literature on judgment, persuasion, and social cognition, and
filtered them against two criteria:
\begin{itemize}
\item \textbf{Text-surface identifiability}: a trigger must produce a linguistic
signal that can be marked as a character span. This rules out biases realized only
in the reader's prior beliefs or mental state rather than in the text. Confirmation
bias, for instance, depends on what a particular reader already believes, so no
span of text carries it on its own.
\item \textbf{Annotation reliability}: candidate biases whose surface signals
annotators could not reliably distinguish from those of another category during
calibration were removed (Section~\ref{sec:annotators}).
\end{itemize}
The resulting schema comprises the nine trigger categories in
Table~\ref{tab:schema}.

\begin{table}[t]
\centering\small
\setlength{\tabcolsep}{4pt}
\begin{tabular}{@{}cl>{\raggedright\arraybackslash}p{3.0cm}@{}}
\toprule
\textbf{ID} & \textbf{Trigger} & \textbf{Ref.} \\
\midrule
\multicolumn{3}{@{}l}{\textit{Credibility}} \\
B01 & Illusory Truth             & \cite{hasher1977frequency} \\
B03 & Bandwagon Effect           & \cite{asch1956studies} \\
B04 & Authority Bias             & \cite{chaiken1980heuristic} \\
\midrule
\multicolumn{3}{@{}l}{\textit{Affect}} \\
B02 & Negativity Bias            & \cite{baumeister2001bad} \\
B07 & Identifiable Victim Effect & \cite{jenni1997explaining} \\
\midrule
\multicolumn{3}{@{}l}{\textit{Social categorization}} \\
B05 & Stereotyping               & \cite{allport1954nature} \\
\midrule
\multicolumn{3}{@{}l}{\textit{Temporal and causal reasoning}} \\
B06 & Declinism                  & \cite{eibach2003change} \\
B08 & Illusory Correlation       & \cite{hamilton1976illusory} \\
B09 & Hindsight Bias             & \cite{fischhoff2003hindsight} \\
\bottomrule
\end{tabular}
\caption{The nine cognitive bias trigger categories in BAIT, grouped by the
mechanism they recruit, with a foundational reference for each.}
\label{tab:schema}
\end{table}

The nine categories draw on four broad mechanisms. Illusory Truth, Bandwagon
Effect, and Authority Bias concern \emph{credibility}: each lends a claim weight
through something other than evidence, whether repetition, popularity, or the
standing of its source. Negativity Bias and the Identifiable Victim Effect
concern \emph{affect}, steering evaluation through emotional salience.
Stereotyping concerns \emph{social categorization}. Declinism, Illusory
Correlation, and Hindsight Bias concern \emph{temporal and causal reasoning}:
how the past is evaluated, how events are linked, and how predictable outcomes
are made to seem. Research on misinformation documents that persuasive and
misleading writing exploits mechanisms of these kinds
\cite{ecker2022psychological}, which makes these triggers relevant to assessing
whether language models can recognize when text is working on its reader.

The nine categories sit at a single level and are non-exclusive: a single
passage may carry several triggers at once, and an article typically contains
instances of several categories. Each category is specified in an annotator
codebook giving its definition, the psychological mechanism it recruits, the
textual conditions that count as a trigger, positive and negative examples, and,
where applicable, its most likely confusion pair. Four such pairs were made explicit during
calibration: Illusory Truth and Negativity Bias (repetition versus asymmetric
weighting), Bandwagon Effect and Authority Bias (the crowd versus a specific
source), Declinism and Hindsight Bias (an evaluative versus an epistemic claim
about the past), and Illusory Correlation and Hindsight Bias (a forward causal
claim versus backward predictability). The full codebook is provided in
Appendix~\ref{app:codebook}. Two rules govern span annotation across all
categories:
\begin{itemize}
\item Under the \emph{minimal-span rule}, a span is the shortest contiguous
fragment that, if removed, would eliminate the trigger, so annotators mark that
fragment rather than the surrounding sentence or paragraph.
\item A single span may also carry more than one trigger, in which case it is
annotated once per applicable category.
\end{itemize}

\section{Dataset Construction}
\label{sec:data}

\subsection{Source Corpus}
\label{sec:source}

We build BAIT on the English subset of the SemEval-2023 Task~3 corpus
\cite{piskorski2023semeval}, 523 news articles on topics including COVID-19,
climate change, migration, and the war in Ukraine. These articles already carry
persuasion-technique and framing labels. We annotate
a random sample of 50 articles, with a mean length of roughly 750 word tokens.

\subsection{Annotators and Training}
\label{sec:annotators}

Two annotators labeled the corpus. The first, a PhD candidate in our research
group, holds an MSc in Psychology. The second, a Master's student in Linguistics
at our university, was recruited through the university's student job platform
and paid the standard university hourly student rate. Before annotating the
released sample, they completed two calibration rounds, during which the schema
was revised and the codebook finalized. A third annotator, the first author of
this paper, then reviewed every label against the finalized codebook and
adjudicated it into the consensus set.

During calibration, several cognitive biases with unreliable surface signals were
removed from an earlier, larger candidate set, explicit confusion pairs were added
to separate categories that annotators tended to conflate, and some definitions
were sharpened. We also fixed the reader's perspective as the unit of judgment:
a trigger is annotated on the basis of the textual cue a passage supplies, not on
any inference about the author's intention.

\subsection{Annotation Protocol}
\label{sec:protocol}

Both annotators labeled each article independently in two passes. The first pass
records, per category, whether the trigger occurs anywhere in the article. The
second marks the minimal span or spans carrying each present trigger, assigns its
category, and records a confidence rating on a three-point scale (1 low,
2 moderate, 3 high).

Confidence governed inclusion in the gold standard. High-confidence triggers were
retained. Moderate- and low-confidence triggers were retained when both
annotators marked them; where the annotators disagreed, the third annotator alone
decided whether to include them. Annotators were instructed to prefer a lower
confidence rating over omitting a suspected trigger, so that borderline cases
survived to adjudication rather than being dropped silently at the annotation
stage.
Spans of a single discontinuous trigger share a group identifier. Annotators did not see the SemEval persuasion or
framing labels, keeping BAIT's labels independent of them.

\subsection{Inter-Annotator Agreement}
\label{sec:iaa}

We measure agreement between the two annotators before adjudication, so the
figures reflect the difficulty of the task rather than agreement with the
consensus labels. Agreement is computed separately for each category, at the
two levels at which annotation took place.

\paragraph{Document level.} In the first annotation pass, each annotator
judges, for each article and category, whether the trigger occurs anywhere in
the article. For each category we compute Cohen's $\kappa$
\cite{cohen1960coefficient} over these binary judgments for the 50 articles,
\begin{equation}
\kappa = \frac{p_o - p_e}{1 - p_e}, \qquad
p_e = p_A\,p_B + (1 - p_A)(1 - p_B),
\label{eq:kappa}
\end{equation}
where $p_o$ is the proportion of articles on which the two annotators agree,
and $p_A$ and $p_B$ are the proportions of articles in which each annotator
marks the category present. The term $p_e$ is the agreement the two annotators
would reach if each marked articles at random at their own rate, so $\kappa$
measures agreement beyond what these rates alone would produce.

\paragraph{Span level.} We split each article into whitespace-delimited tokens
and, for each annotator and category, mark the tokens that overlap at least one
of that annotator's spans. With $A$ and $B$ the sets of tokens marked by the two
annotators, pooled over all 50 articles, we compute
\begin{equation}
F_1 = \frac{2\,|A \cap B|}{|A| + |B|}.
\label{eq:iaa-f1}
\end{equation}
This measure is symmetric, so neither annotator needs to be treated as the
reference. We use token $F_1$ rather than $\kappa$ because free-form span
selection has no fixed set of negative items over which chance agreement could
be defined \cite{hripcsak2005agreement}. This measure does not consider how spans are grouped
into triggers. Averages are unweighted means over the nine categories.

\begin{table}[t]
\centering\small
\begin{tabular}{@{}clcc@{}}
\toprule
\textbf{ID} & \textbf{Trigger} & \textbf{$\kappa$ (doc)} & \textbf{$F_1$ (span)} \\
\midrule
B01 & Illusory Truth             & 0.55 & 0.62 \\
B02 & Negativity Bias            & 0.47 & 0.48 \\
B03 & Bandwagon Effect           & 0.23 & 0.46 \\
B04 & Authority Bias             & 0.56 & 0.46 \\
B05 & Stereotyping               & 0.53 & 0.46 \\
B06 & Declinism                  & 0.56 & 0.66 \\
B07 & Identifiable Victim Effect & 0.62 & 0.55 \\
B08 & Illusory Correlation       & 0.60 & 0.50 \\
B09 & Hindsight Bias             & 0.63 & 0.33 \\
\midrule
    & \textbf{Average}           & \textbf{0.53} & \textbf{0.50} \\
\bottomrule
\end{tabular}
\caption{Per-category inter-annotator agreement before adjudication.
Document-level Cohen's $\kappa$ on trigger presence and span-level
word-token $F_1$, both over the 50 articles.}
\label{tab:iaa}
\end{table}

Table~\ref{tab:iaa} reports per-category agreement. Averaged over the nine
categories, agreement is $\kappa = 0.53$ at the document level and $F_1 = 0.50$
at the span level. Document-level $\kappa$ ranges from $0.23$ (Bandwagon
Effect) to $0.63$ (Hindsight Bias), and span-level $F_1$ from $0.33$
(Hindsight Bias) to $0.66$ (Declinism). The lowest $\kappa$ belongs to the
rarest category. The annotators marked Bandwagon Effect in only three and four
of the 50 articles respectively, so chance agreement is high ($p_e = 0.87$) and
their raw agreement on 45 of 50 articles yields a $\kappa$ of only $0.23$. This
sensitivity of $\kappa$ to rare categories is well documented
\cite{feinstein1990high}. Negativity Bias shows a different pattern: one
annotator marked it in 24 articles and the other in 11, so its disagreements
reflect a difference in threshold rather than scattered judgments.

We acknowledge that this agreement is moderate, and argue that it is the
expected outcome rather than a sign of weak annotation. Each category is
defined by the cognitive mechanism a passage sets off in a reader, not by a
fixed surface form that annotators could match word for word. Two trained
annotators reading the same passage can therefore disagree in good faith over
whether it supplies the cue. We take the agreement they reach as the practical
ceiling for the task, the level careful human annotators attain, and evaluate
models against that level.

\subsection{Dataset Statistics}
\label{sec:stats}

The consensus set holds 592 trigger spans, grouped into 398 triggers, across
the 50 articles. Forty-six articles carry at least one trigger, and four carry
none and are labeled \textsc{none}, giving negative examples for detection.
Table~\ref{tab:distribution} reports per-category coverage.

\begin{table}[t]
\centering
\small
\setlength{\tabcolsep}{2.4pt}
\begin{tabular}{@{}llrrrr@{}}
\toprule
ID & Bias & Art. & Spans & \% & Len. \\
\midrule
B01 & Illusory Truth             & 45 & 338 & 57.1 & 14.6 \\
B02 & Negativity Bias            & 23 &  86 & 14.5 & 10.3 \\
B03 & Bandwagon Effect           &  7 &  11 &  1.9 & 10.3 \\
B04 & Authority Bias             & 19 &  34 &  5.7 & 28.1 \\
B05 & Stereotyping               & 14 &  25 &  4.2 & 17.9 \\
B06 & Declinism                  &  6 &  17 &  2.9 & 35.0 \\
B07 & Identifiable Victim Effect &  7 &  20 &  3.4 & 26.4 \\
B08 & Illusory Correlation       & 16 &  42 &  7.1 & 30.0 \\
B09 & Hindsight Bias             & 11 &  19 &  3.2 & 13.4 \\
\midrule
\multicolumn{2}{@{}l}{Total} & 46 & 592 & 100 & 16.9 \\
\bottomrule
\end{tabular}
\caption{BAIT label distribution over 50 articles. \emph{Art.}: articles
containing the bias (the total counts articles with at least one bias).
\emph{Spans}: annotated spans. \emph{\%}: share of all 592 spans.
\emph{Len.}: mean span length in words.}
\label{tab:distribution}
\end{table}

\subsection{Relation to Persuasion Techniques}
\label{sec:semeval}

\begin{figure}[t]
\centering
\includegraphics[width=\linewidth]{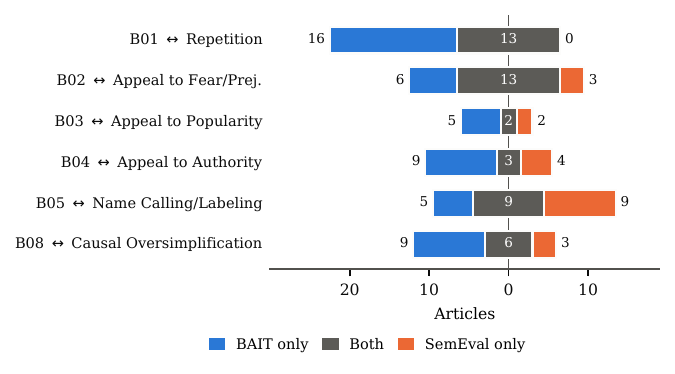}
\caption{Article-level overlap between BAIT categories and their closest
SemEval-2023 persuasion technique, on the 32 articles with at least one SemEval
label: articles labeled with the BAIT category only, with both, and with the
technique only. Declinism, the Identifiable Victim Effect and Hindsight Bias
have no counterpart.}
\label{fig:semeval}
\end{figure}

Because BAIT is built on the SemEval-2023 Task~3 corpus, its labels can be
compared with the persuasion techniques annotated for that task
\cite{piskorski2023semeval}. These techniques are available to us only at the
article level, so we compare at that level, on the 32 articles that carry at
least one technique.

Six BAIT categories have a natural counterpart among the
techniques (Figure~\ref{fig:semeval}). Agreement between each category and its
counterpart is low, with Cohen's $\kappa$ between 0.05 and 0.24, except for
Negativity Bias and Appeal to Fear/Prejudice ($\kappa = 0.44$). The differences
follow from BAIT's definitions. Every article labeled with Repetition also
carries Illusory Truth, but Illusory Truth occurs in 16 further articles,
because BAIT also counts reworded restatements of a claim. Of the 12 of these
articles with Authority Bias, only 3 carry Appeal to Authority, because BAIT
requires a credential that replaces evidence rather than any appeal to a
source. Three BAIT categories have no counterpart at all and account for 18\%
of BAIT's article-level labels on these articles.
BAIT therefore does not duplicate the existing label set: describing what a
passage does to a reader captures distinctions that a taxonomy of persuasion
techniques does not.

\section{Tasks and Metrics}
\label{sec:tasks}

We formulate bias-trigger annotation as two tasks of increasing difficulty:
detecting which biases an article contains, and localizing and grouping the
spans that trigger them. Both operate on a full news article and target the
nine categories (B01--B09), and we evaluate each under zero-shot and few-shot
prompting.

\paragraph{Scoring.} Every metric below compares a model's output with the
gold standard unit by unit, where the unit depends on the metric: a
(document, category) decision, a character, a span, or a trigger. For each
category $b$, a true positive ($\mathrm{TP}_b$) is a predicted unit that
matches a gold unit, a false positive ($\mathrm{FP}_b$) is a predicted unit
that matches none, and a false negative ($\mathrm{FN}_b$) is a gold unit that
no prediction matches. We sum these counts over all 50 articles and all nine
categories and compute a single micro-averaged $F_1$ for the corpus,
\begin{equation}
F_1 = \frac{2\sum_b \mathrm{TP}_b}
  {2\sum_b \mathrm{TP}_b + \sum_b \mathrm{FP}_b + \sum_b \mathrm{FN}_b}.
\label{eq:f1-micro}
\end{equation}

\subsection{Task 1: Bias Trigger Detection}

Given an article, the model identifies which bias categories are present. This
is multi-label classification over the nine categories, and an article may
exhibit any subset of biases, including none (4 of the 50 articles contain no
trigger). The task measures whether a model recognizes that an article employs
a given bias, independently of where.

The unit is a (document, category) decision. A category the model predicts for
an article is a true positive if the gold standard also lists it for that
article and a false positive otherwise, and a gold category the model misses is
a false negative. We report $F_1$ (Equation~\ref{eq:f1-micro}) together with
subset accuracy, the fraction of the $N$ articles whose predicted category set
$\hat{Y}_d$ exactly matches the gold set $Y_d$,
\begin{equation}
\mathrm{SubsetAcc} = \frac{1}{N} \sum_{d=1}^{N}
  \mathbf{1}\bigl[\hat{Y}_d = Y_d\bigr].
\label{eq:subset-acc}
\end{equation}
The gold category set of an article is exactly the set of categories with at
least one gold trigger, so a model is not penalized for answering the two tasks
consistently.

\subsection{Task 2: Bias Trigger Localization}

Given an article, the model extracts the spans that trigger each bias and
groups them into trigger instances. This reflects BAIT's central annotation
decision that a single bias instance may be realized across several
non-contiguous spans. Illusory-truth effects (B01), for example, arise from a
claim repeated through an article, so the repeated mentions form one trigger of
several spans rather than several independent triggers. Task~2 therefore
requires both locating biased text and correctly partitioning spans into
triggers. In our corpus, 143 of 398 gold triggers (36\%) consist of more than
one span.

As a human reference for both tasks, we score one annotator's labels against
the other's, before adjudication, with exactly the metrics used for the models.
Because every metric here is symmetric, the choice of reference annotator does
not matter. We do not score the annotators against the gold standard, which was
built from their own labels.

\paragraph{Locating spans.} A span is a contiguous stretch of characters in the
article, identified by its first and last character. Because models return
spans as copied strings, we first locate each string in the article, trying
four matchers in turn: an exact match, a case-insensitive match, a match that
tolerates differences in whitespace, and a fuzzy match that accepts a window of
the article with a character similarity of at least 0.80 to the string. When a
string occurs several times, as repeated claims often do, we prefer an
occurrence not yet used by another span of the same category in the same
output, then one that overlaps a gold span of that category, then the
leftmost. The gold standard thus only chooses between positions where the
article contains exactly the returned string, and cannot turn a wrong string
into a correct one. At least 95\% of every model's spans match exactly, and at
most 0.4\% cannot be located at all (Appendix~\ref{app:alignment}). A string
that cannot be located is \emph{unlocated} and counts against the model at
every level below. We score localization at three levels of granularity, from
most to least forgiving, each reported as $F_1$ (Equation~\ref{eq:f1-micro}).

\paragraph{Character level.} The unit is a character of the article. For each
category, a character is a true positive if it lies inside both a predicted
and a gold span, a false positive if it lies only inside a predicted span, and
a false negative if it lies only inside a gold span. The characters of an
unlocated span all count as false positives. This rewards partial overlap, in
the spirit of the character-overlap scoring used for propaganda span
identification \cite{da-san-martino-etal-2020-semeval}.

\paragraph{Span level.} The unit is a span. We measure the overlap of a
predicted span $a$ and a gold span $g$ by their intersection over union,
\begin{equation}
\mathrm{IoU}(a, g) = \frac{|a \cap g|}{|a \cup g|},
\label{eq:iou}
\end{equation}
where $|\cdot|$ counts characters. This is the Jaccard index
\cite{jaccard1912distribution}, known as IoU in object-detection evaluation
\cite{everingham2010pascal}. It is 1 when the two spans coincide and 0 when they
do not overlap. Predicted and gold spans of the same category are paired
one-to-one, taking the pairs with the highest IoU first, and a pair is a true
positive when its IoU is at least $\tau_s = 0.5$ (at this threshold the greedy
pairing is optimal; Appendix~\ref{app:scores}). Unpaired predicted spans,
including unlocated ones, are false positives, and unpaired gold spans are
false negatives. This penalizes boundary errors that the character level partly
forgives.

\paragraph{Trigger level.} The unit is a trigger, the set of spans that the
model or the annotator assigns to one bias instance. For a predicted trigger
$\hat{T}$ and a gold trigger $T$, let $C(\hat{T})$ and $C(T)$ be the sets of
characters covered by their spans. We score their agreement by the character
overlap
\begin{equation}
F(\hat{T}, T) = \frac{2\,\bigl|C(\hat{T}) \cap C(T)\bigr|}
                     {\bigl|C(\hat{T})\bigr| + \bigl|C(T)\bigr|},
\label{eq:trigger-f1}
\end{equation}
where the length of any unlocated span in $\hat{T}$ is added to
$|C(\hat{T})|$. Predicted and gold triggers of the same category are paired
one-to-one in decreasing order of $F$, and a pair is a true positive when
$F(\hat{T}, T) \geq \tau_t = 0.5$. Unpaired predicted and gold triggers are
false positives and false negatives respectively. This is the only metric that
evaluates grouping. A model that locates every span correctly but splits one
multi-span gold trigger into several single-span triggers can score well at the
character and span levels, but is penalized here.

\begin{table*}[t]
\centering\small
\setlength{\tabcolsep}{5pt}
\begin{tabular}{llcccccc}
\toprule
 & & \multicolumn{3}{c}{Task 1: detection} & \multicolumn{3}{c}{Task 2: localization} \\
\cmidrule(lr){3-5}\cmidrule(lr){6-8}
Model & $k$ & Micro-F1 & Macro-F1 & Subset Acc. & Char-F1 & Span-F1 & Trigger-F1 \\
\midrule
\multirow{3}{*}{Claude Opus 4.8}
 & 0  & 0.543 & 0.478 & 0.040 & 0.147 & 0.121 & 0.102 \\
 & 3  & 0.637 & \textbf{0.563} & 0.060 & 0.201 & 0.167 & 0.157 \\
 & 10 & 0.679 & 0.590 & 0.080 & 0.234 & 0.204 & 0.179 \\
\midrule
\multirow{3}{*}{GPT-5.6}
 & 0  & 0.590 & 0.491 & 0.033 & 0.209 & 0.165 & 0.150 \\
 & 3  & 0.614 & 0.531 & \textbf{0.073} & 0.241 & 0.204 & 0.180 \\
 & 10 & 0.643 & 0.559 & 0.087 & 0.263 & 0.232 & 0.216 \\
\midrule
\multirow{3}{*}{Grok 4.5}
 & 0  & \textbf{0.625} & \textbf{0.560} & 0.020 & 0.216 & \textbf{0.188} & \textbf{0.153} \\
 & 3  & \textbf{0.653} & 0.543 & 0.067 & \textbf{0.269} & \textbf{0.227} & \textbf{0.193} \\
 & 10 & 0.680 & \textbf{0.598} & 0.087 & \textbf{0.275} & \textbf{0.239} & \textbf{0.223} \\
\midrule
\multirow{3}{*}{DeepSeek-V4}
 & 0  & 0.448 & 0.378 & 0.000 & 0.093 & 0.064 & 0.060 \\
 & 3  & 0.543 & 0.478 & 0.013 & 0.149 & 0.107 & 0.123 \\
 & 10 & 0.654 & 0.536 & 0.053 & 0.228 & 0.163 & 0.178 \\
\midrule
\multirow{3}{*}{Kimi K3}
 & 0  & 0.601 & 0.516 & \textbf{0.067} & 0.193 & 0.166 & 0.144 \\
 & 3  & 0.641 & 0.554 & 0.067 & 0.225 & 0.205 & 0.184 \\
 & 10 & \textbf{0.685} & 0.596 & \textbf{0.127} & 0.251 & 0.216 & 0.212 \\
\midrule
\multirow{3}{*}{Mistral Medium 3.5}
 & 0  & 0.582 & 0.518 & 0.003 & \textbf{0.228} & 0.155 & 0.130 \\
 & 3  & 0.613 & 0.531 & 0.040 & 0.230 & 0.161 & 0.147 \\
 & 10 & 0.636 & 0.562 & 0.047 & 0.240 & 0.181 & 0.170 \\
\midrule
\textit{Annotator vs.\ annotator} & -- & \textit{0.745} & \textit{0.642} & \textit{0.360} & \textit{0.569} & \textit{0.493} & \textit{0.502} \\
\bottomrule
\end{tabular}
\caption{Detection (Task~1) and localization (Task~2) for $k$ in-context
examples. Corpus-level scores over 50 articles. For $k>0$, the mean over three
example draws (seeds 17, 23, 29); for $k=0$, the mean over six zero-shot runs.
Bold: best model at each $k$. The last row scores one annotator's labels
against the other's before adjudication, with the same metrics, as a human
reference.}
\label{tab:main}
\end{table*}

\begin{figure*}[t]
\centering
\includegraphics[width=0.9\linewidth]{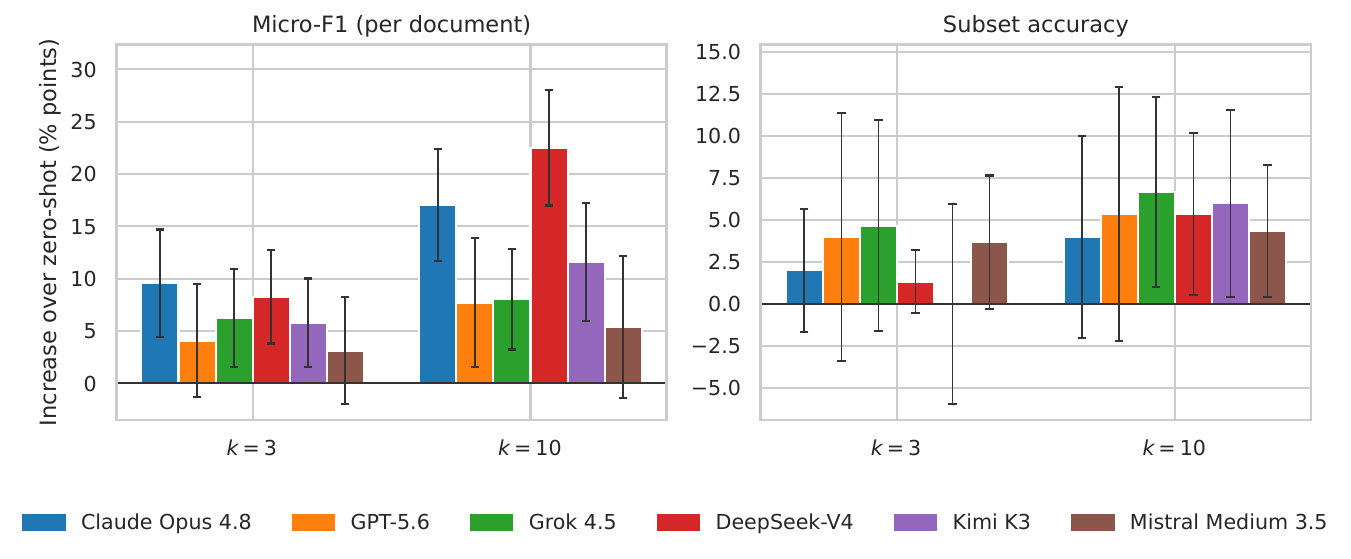}
\caption{Task 1: increase over zero-shot when the model receives $k=3$ or
$k=10$ annotated example articles, in percentage points. Bars show the mean
per-article difference (few-shot averaged over three example draws, zero-shot
over six runs), and error bars are paired 95\% confidence intervals over the 50
articles. Left: micro-F1. Right: subset accuracy (the predicted set of biases
equals the gold set).}
\label{fig:task1}
\end{figure*}

\begin{figure*}[t]
\centering
\includegraphics[width=\linewidth]{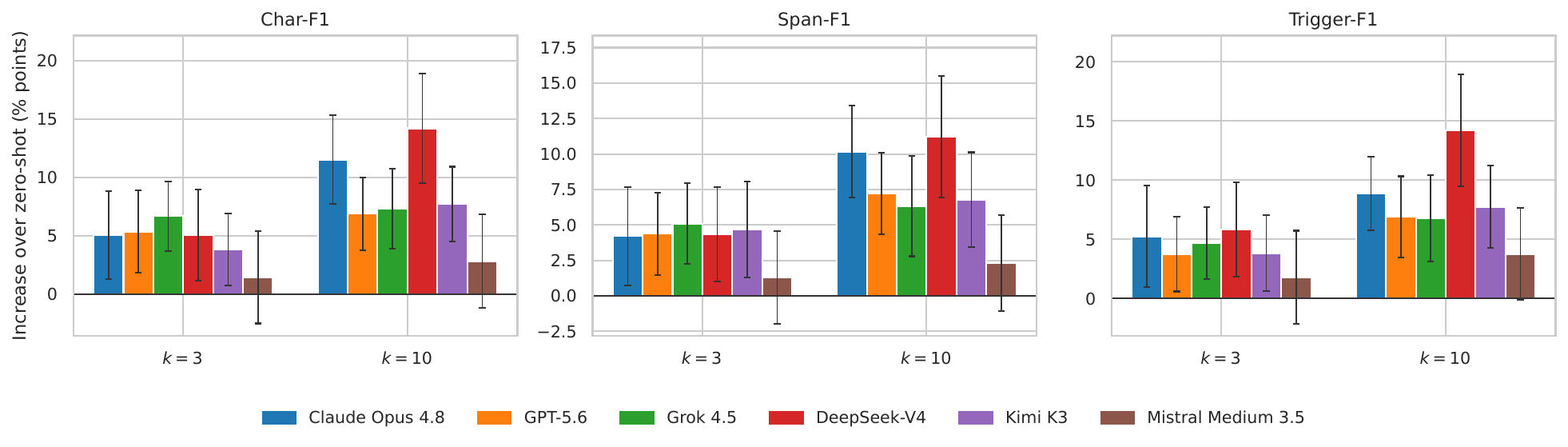}
\caption{Task 2: increase over zero-shot at three levels of strictness, in percentage points. Bars show the mean per-article difference (few-shot averaged over three example draws, zero-shot over six runs); error bars are 95\% confidence intervals (paired $t$-interval over the 50 articles). Zero-shot levels are in Table~\ref{tab:main}.}
\label{fig:task2}
\end{figure*}

\section{Experimental Setup}
\label{sec:setup}

 We evaluate six frontier models, Claude Opus 4.8, GPT-5.6, Grok 4.5, DeepSeek-V4,
Kimi K3, and Mistral Medium 3.5, at three in-context settings. Every model receives the same
system prompt in every setting. It describes the two tasks, gives a one-line
definition of each bias, states the rule for multi-span triggers, and fixes
the output format.

In the zero-shot
setting ($k=0$) the model receives only the instructions. In the few-shot settings
it additionally receives $k=3$ or $k=10$ fully annotated example articles, drawn
from the rest of the corpus under a leave-one-out protocol so that no article
appears among its own examples. Each few-shot setting is run with three random
draws of the examples (seeds 17, 23, and 29), and we report the mean over the
three draws. The system prompt, bias definitions, and output format are identical
across settings, so the settings differ only in the in-context examples.

Because outputs varied across repeated queries, even with temperature set to 0,
zero-shot scores are the mean over six runs, one accompanying each few-shot run. The
article is the sampling unit, so we compare settings and models with paired
$t$-tests over the 50 articles on per-article scores, Holm-corrected
\cite{holm1979simple} for the number of comparisons in each figure. Outputs that
cannot be parsed are scored as empty predictions rather than excluded, so every
model and setting is evaluated on the same 50 articles. Decoding settings and
scoring details are in Appendices~\ref{app:models} and~\ref{app:scores}.

\section{Results}
\label{sec:results}

\begin{figure*}[t]
\centering
\includegraphics[width=0.72\linewidth]{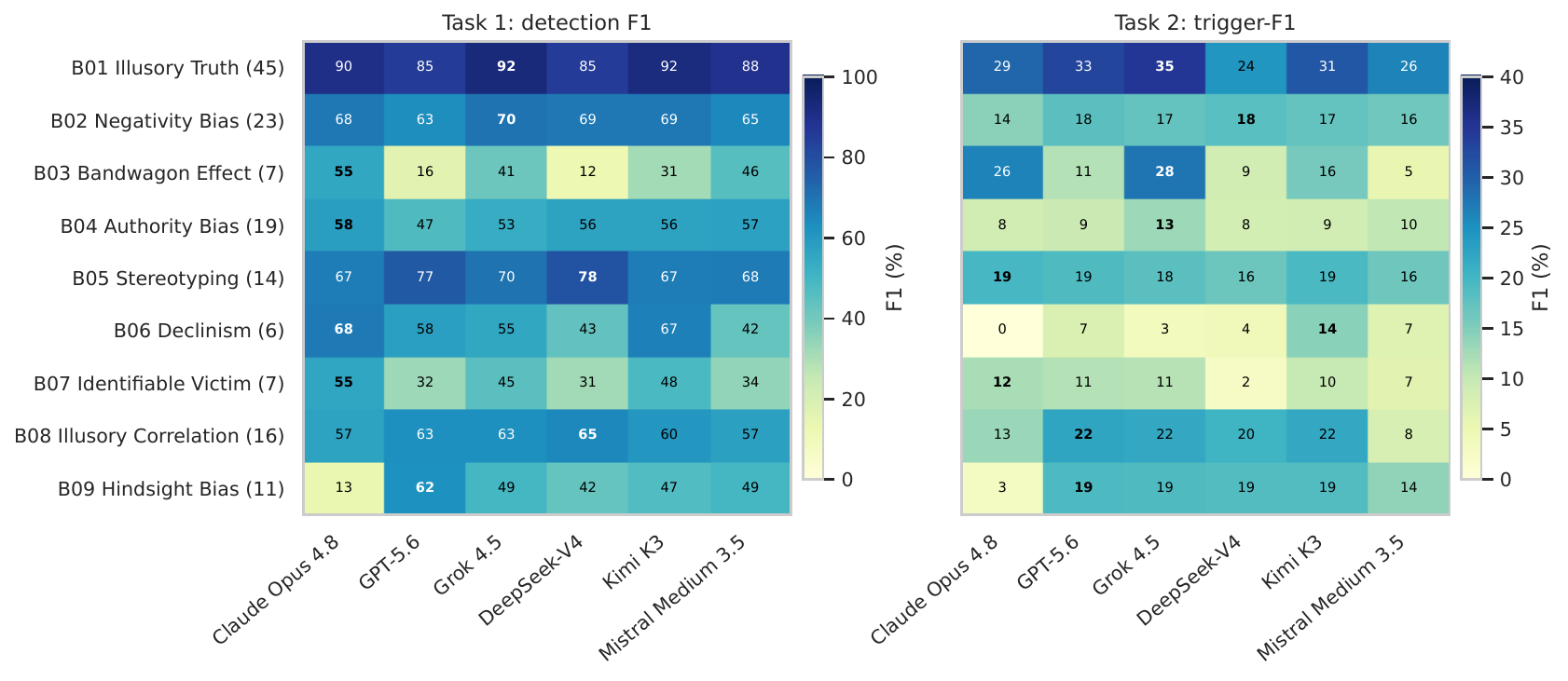}
\caption{Per-bias scores with ten in-context examples (corpus-level $F_1$ in \%,
mean over three example draws). Left: Task~1 detection $F_1$ (scale 0--100).
Right: Task~2 trigger-F1 (scale 0--40). Bold: best model on that bias for that
task. Number of gold articles per bias in parentheses.}
\label{fig:bias}
\end{figure*}

Table~\ref{tab:main} reports corpus-level scores for all six models at $k \in \{0, 3, 10\}$. Figures~\ref{fig:task1} and~\ref{fig:task2} show how much the examples add over zero-shot on each task, and Figure~\ref{fig:bias} breaks both tasks down by bias. We call a difference significant only if it survives Holm correction.

\begin{figure*}[t]
\centering
\includegraphics[width=0.72\linewidth]{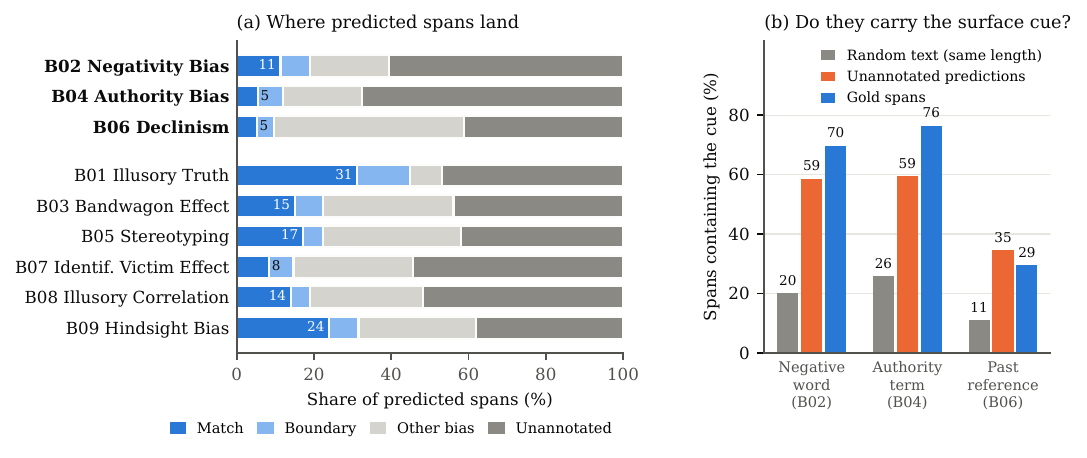}
\caption{Where predicted spans land, and whether they carry the surface cue
($k=10$, all six models and three example draws, 11{,}027 located spans).
(a) Each predicted span is classified as a \emph{match} (IoU $\geq 0.5$ with a
gold span of the same bias), a \emph{boundary} error (overlapping such a span
with lower IoU), an overlap with a gold trigger of \emph{another bias}, or
\emph{unannotated} text that no annotator marked for any bias. Numbers give the
match share in \%. (b) Share of spans containing the bias's surface cue, for
unannotated predictions, gold spans, and random stretches of article text
matched to the predictions in length.}
\label{fig:cue}
\end{figure*}

\subsection{Detecting Bias Triggers (Task 1)}
\label{sec:res-task1}

Zero-shot, detection micro-F1 ranges from 0.45 (DeepSeek-V4) to 0.62
(Grok~4.5). With ten examples the models converge on 0.64--0.69, led by
Kimi~K3 (0.685), Grok~4.5 (0.680) and Claude Opus~4.8 (0.679). The two
annotators reach 0.745 against each other (Table~\ref{tab:main}), so the best
models come within about six points of human agreement.

Figure~\ref{fig:task1} shows that the benefit of examples depends strongly on
the model. The two weakest zero-shot models gain the most: with ten examples,
DeepSeek-V4 improves by 22.5 points per article and Claude Opus~4.8 by 17.1.
Kimi~K3 (+11.6) and Grok~4.5 (+8.0) also improve significantly, whereas the
gains of GPT-5.6 (+7.7) and Mistral Medium~3.5 (+5.4) do not survive
correction. Three examples are not enough for most models. At $k=3$, only
Claude Opus~4.8 (+9.5) and DeepSeek-V4 (+8.3) improve significantly.

Examples do not, however, help models reproduce an article's complete bias
profile. Subset accuracy is at most 0.13 for every model and setting, against
0.36 between the two annotators, and none of its gains is significant.

\textbf{Key Finding 1.} \emph{LLMs approach human agreement in recognizing
which biases an article contains, but rarely recover its full set of biases,
and in-context examples mainly lift the weakest models toward the others
rather than closing this gap.} A model that screens news for manipulative
framing would flag many manipulative articles while often misreporting which
mechanisms they rely on.

 \subsection{Localizing Bias Triggers (Task 2)}
\label{sec:res-task2}

Table~\ref{tab:main} shows that localization is much harder than detection.
Zero-shot, trigger-F1 ranges from 0.06 (DeepSeek-V4) to 0.15 (Grok~4.5 and
GPT-5.6), and with ten examples from 0.17 to 0.22, led by Grok~4.5 (0.223). At
$k=10$, every model's detection micro-F1 is 3.0 to 3.8 times its trigger-F1.
The gap to human agreement is far larger than for detection. The two
annotators reach a trigger-F1 of 0.50 against each other, more than twice the
best model's 0.22, and the same holds at the character level (0.57 against at
most 0.28).

Examples help localization more consistently than detection (Figure~\ref{fig:task2}). With ten examples, five of the six models improve significantly at every granularity. On trigger-F1, the gains are 14.2 points for DeepSeek-V4, 8.9 for Claude Opus~4.8, 7.7 for Kimi~K3, 6.9 for GPT-5.6 and 6.8 for Grok~4.5. Mistral Medium~3.5 is the exception: its gains (2.3--3.8 points) are not significant at any granularity. Three examples are not enough. At $k=3$, no model improves significantly on trigger-F1, and only Grok~4.5 improves significantly at the character and span levels. These gains are averages over articles and hide variation between them: with ten examples, trigger-F1 improves on 28--36 articles per model but gets worse on 4--13.

\textbf{Key Finding 2.} \emph{LLMs largely fail to show where a bias trigger
is. The best model reaches less than half the trigger-level agreement of two
human annotators (0.22 against 0.50), and examples improve this only
modestly.}
\subsection{Differences Across Biases}
\label{sec:res-bias}

The aggregate scores hide large differences between biases
(Figure~\ref{fig:bias}). Averaged over models, detection $F_1$ ranges from 0.89
on Illusory Truth (B01) to 0.34 on Bandwagon Effect (B03). The high score on
Illusory Truth largely reflects its prevalence: it occurs in 45 of the 50
articles, so a model that predicts it almost everywhere is rarely wrong.
Localization is uniformly poor. Illusory Truth is localized best (mean
trigger-F1 0.30), followed by Stereotyping (B05, 0.18) and Illusory Correlation
(B08, 0.18). Declinism (B06, 0.06), the Identifiable Victim Effect (B07, 0.09)
and Authority Bias (B04, 0.10) are localized worst. Relative to detection, the
loss is also largest for these three: detection $F_1$ is 4.6 to 9.2 times their
trigger-F1, against 2.1 to 4.0 times for the other six biases. Authority Bias,
for instance, falls from 0.55 to 0.10 and Declinism from 0.55 to 0.06. The
annotators, by contrast, agree on Authority Bias spans at a token-level $F_1$
of 0.46 and on Declinism spans at 0.66 (Table~\ref{tab:iaa}), so these
passages are not inherently hard to delimit.

\paragraph{Excluding Illusory Truth.} Because Illusory Truth accounts for 57\% of gold spans and almost all
multi-span triggers, we check that it does not drive the main results. Without
it, detection micro-F1 at $k=10$ ranges from 0.55 to 0.60 against 0.64 between
the annotators, and macro-F1 from 0.50 to 0.56 against 0.61. Localization falls
further: trigger-F1 ranges from 0.12 to 0.16, against 0.44 between the
annotators. Illusory Truth therefore flatters the models. On the remaining
eight biases, they stay close to human agreement on detection but reach only
about a third of it on localization.

The models' errors point to an explanation. With ten examples, models predict
Negativity Bias (B02) in 39 articles on average against 23 in the gold
standard, Authority Bias in 30 against 19, and Declinism in 10.5 against 6. All
three have an easy surface cue (negative vocabulary, a cited expert, a
reference to the past) that the codebook explicitly does not accept as
sufficient. It requires \emph{asymmetric} negative weighting, a credential that
\emph{replaces} evidence, and an idealized past whose problems are omitted.
Hindsight Bias (B09) shows the opposite failure. It is under-applied, predicted
in 7.5 articles against 11, and Claude Opus~4.8 almost never predicts it (in
1.7 articles on average), which explains its detection $F_1$ of 0.13 on this
bias against 0.42 to 0.62 for the other models.

No model is best across the board. Claude Opus~4.8 has the best detection $F_1$
on four biases, but the worst on Hindsight Bias and zero trigger-F1 on
Declinism. Grok~4.5 localizes three biases best, and Mistral Medium~3.5 is
weakest at localizing Bandwagon Effect and Illusory Correlation. With as few as
six gold articles for some biases, most of these per-bias differences are
within noise. Across whole tasks, comparing every pair of models with paired
$t$-tests over articles (Holm-corrected over the 15 pairs), only two pairs
differ significantly at $k=10$: Grok~4.5 and Kimi~K3 each localize better than
Mistral Medium~3.5, by 5.6 and 6.5 points of trigger-F1.

\textbf{Key Finding 3.} \emph{Model performance depends on the bias more than
on the model.} Detection ranges from 0.89 on Illusory Truth to 0.34 on
Bandwagon Effect, and localization is weakest for Declinism, the Identifiable
Victim Effect and Authority Bias. Models over-predict the three biases with the
most recognizable cues (Negativity Bias, Authority Bias and Declinism) and
under-predict Hindsight Bias, while differences between models on any single
bias are mostly within noise.

\subsection{Surface Cues Versus Triggers}
\label{sec:res-cues}

Section~\ref{sec:res-bias} showed that models over-predict the three biases
with the easiest surface cues. We test whether this reflects attention to the
cue rather than to the mechanism by examining where the predicted spans land
(Figure~\ref{fig:cue}a). Of all spans predicted at $k=10$ for Negativity Bias,
Authority Bias and Declinism, only 9.2\% match a gold span of the same bias,
against 26.9\% for the other six biases, and 60.8\% fall on text that no
annotator marked for any bias. For Authority Bias, 53\% of predicted spans lie
in articles that contain no Authority Bias trigger at all. The pattern holds
for every model, with 8 to 11\% of spans matched for these three biases,
against 21 to 32\% for the others.

These misplaced spans do carry the cue. We check each one for a simple lexical
marker: a negative word from the VADER lexicon \cite{hutto2014vader} for
Negativity Bias; an authority term (e.g.\ ``expert'', ``study'',
``according to'', a title or an institution) for Authority Bias; and a
reference to the past (e.g.\ a year, ``used to'', ``no longer'') for
Declinism. As a baseline, we draw five random stretches of article text of the
same length for each prediction. The misplaced spans contain the cue far more
often than chance (Figure~\ref{fig:cue}b): 59\% of unannotated Negativity Bias
predictions contain a negative word, 59\% of Authority Bias predictions an
authority term, and 35\% of Declinism predictions a reference to the past,
against 20\%, 26\% and 11\% for random text.

The four articles without any gold trigger illustrate the pattern. In one of
them, all six models mark a sentence attributing a factual claim to a public
health agency as Authority Bias, and a plain report of a death toll as
Negativity Bias. Neither passage meets the codebook's conditions: the agency is
cited as the source of a fact, not as a substitute for evidence, and the death
toll is stated without elaboration or asymmetric weighting. Each passage
carries the surface cue of its bias, a named institution or a negative fact,
but not the mechanism.

\textbf{Key Finding 4.} \emph{Models respond to the surface cue of a bias
rather than to its mechanism.} Most of the spans they mark for Negativity Bias,
Authority Bias and Declinism fall on text that no annotator marked, yet these
spans contain the bias's surface cue two to three times as often as random text.
A bias detector built on current LLMs would thus flag these biases most readily
where it is least able to show them.

\section{Limitations and Future Work}
\label{sec:limitations}

\paragraph{Scale and coverage.} BAIT covers 50 English news articles from a
single source corpus, on a limited set of topics (COVID-19, climate change,
migration, and the war in Ukraine). Because the sample is random, category
frequencies follow the source corpus rather than a balanced design. Illusory
Truth occurs in 45 articles, whereas the sparsest categories rest on six or
seven, so per-bias scores for these categories are imprecise, and most
per-bias differences between models fall within noise. Pooled scores are also
dominated by the frequent categories, which is why we report per-bias results
separately. Whether the findings hold for other languages, genres, and outlets
remains to be tested.

\paragraph{Annotation.} Each category is defined by the effect a passage is
likely to have on a reader, so annotation requires judgment, and agreement
between the two annotators is moderate ($\kappa = 0.53$, token $F_1 = 0.50$).
The two annotators bring training in psychology and in linguistics, and a
single adjudicator, who is also an author of this paper, resolved their
labels into the gold standard. A larger and more varied pool of annotators,
with independent adjudication, would give a more robust consensus and a better
estimate of human agreement on each category. Triggers are also annotated for
a generic reader. Whether a passage actually biases a particular person
depends on what that person already knows and believes, and BAIT marks
passages likely to trigger a bias, not measured effects on readers.

\paragraph{Evaluation.} We evaluate six models through prompting alone, with
one-line definitions of each bias rather than the full codebook used by the
annotators, and with up to ten in-context examples. The models therefore
received less guidance than the annotators did, and our results describe what
current models do under this setup, not an upper bound on what they can learn.
Model behavior is also not fully reproducible: outputs varied across repeated
queries even with temperature set to 0, which we address by averaging over
runs, and API-served models may change over time.

\paragraph{Data release.} The articles belong to the SemEval-2023 Task~3
corpus, whose terms do not permit redistribution. We therefore release BAIT as
standoff annotations, with article identifiers and character offsets, so that
users with access to the source corpus can reconstruct the full dataset.

\paragraph{Future work.} Three directions follow directly from these
limitations. First, extending the annotation to more articles, languages and
outlets would yield enough instances of the sparse categories to compare models
on each bias reliably. Second, our results raise the question of whether the
gap between detection and localization reflects missing knowledge or missing
guidance. Giving models the full codebook, fine-tuning them on the gold
spans, or asking them to justify each trigger against the codebook's conditions
would test whether they can learn to look past the surface cue to the
mechanism. Third, conditioning trigger detection on an explicit model of the
reader, and testing annotated triggers against reader responses in user
studies, would connect BAIT's labels to the effects they are meant to
anticipate.
\section{Conclusion}
\label{sec:conclusion}

We introduced BAIT, a resource for detecting cognitive bias triggers in news:
the features of a text that recruit a reader's mental shortcuts to shape
belief. BAIT pairs a schema of nine triggers, each specified in an annotator
codebook, with article- and span-level annotations of 50 English news articles
adjudicated into a gold standard. Its labels describe what a passage does to a
reader, and they overlap only weakly with the persuasion techniques already
annotated on the same articles. On this data we define two tasks, detecting
which triggers an article contains and locating the spans that carry them.

Benchmarking six frontier LLMs on these tasks shows that models can tell that
an article is manipulative, but not how or where. With ten in-context
examples, detection micro-F1 reaches 0.64--0.69, close to the 0.75 that the two
annotators reach with each other, yet no model recovers an article's full set
of biases in more than 13\% of articles, against 36\% for the annotators.
Localization is much weaker: the best trigger-F1 is 0.22, less than half the
trigger-F1 of 0.50 that the annotators reach with each other. In-context
examples help most for the weakest models but do not close these gaps.

The errors point to a common cause. Models over-apply the biases with the
easiest surface cues: negative vocabulary, a cited expert, a reference to the
past. Most of the spans they mark for these biases fall on text that no
annotator marked, yet these spans contain the cue two to three times as often
as random text of the same length. Current LLMs thus respond to what a trigger
looks like rather than to the conditions that make a passage one.

This matters for any system that relies on an LLM to screen, summarize, or
rank news on a reader's behalf. Such a system would flag many manipulative
articles, but it would often misreport which mechanisms they rely on and could
not point to the passages responsible, so its judgments could not be checked
by the people they are meant to protect. We release the BAIT schema, codebook,
standoff annotations, and evaluation pipeline to support work on models that
can show, and not only state, why a text may bias its reader.

\section*{Acknowledgments}
We would like to thank our annotators, No\^{e}mie De Vlaminck and Yumi
Slingeneyer, for their careful work and for their thoughtful contributions
during the calibration rounds, which shaped the final schema and codebook.
We also thank our colleague Liesbeth Allein, and Els Lefever, Aaron Maladry
and Lo\"{i}c De Langhe of the Language and Translation Technology Team (LT3)
at Ghent University, for their feedback on this work and their help in recruiting one of our annotators. The research leading to these results was funded/co-funded by the European Union (ERC, VIGILIA, 101142229), the Special Research Fund (BOF) of Ghent University (BOF20/IBF/117), the Flemish Government under the ``Onderzoeksprogramma Artificiële Intelligentie (AI) Vlaanderen'' programme, and the FWO (project no. G073924N). Views and opinions expressed are however those of the author(s) only and do not necessarily reflect those of the European Union or the European Research Council Executive Agency. Neither the European Union nor the granting authority can be held responsible for them. For the purpose of Open Access the author has applied a CC BY public copyright license to any Author Accepted Manuscript version arising from this submission.

\bibliography{aaai2027}

\appendix

\section{Models}
\label{app:models}

We evaluated six frontier large language models.
\begin{itemize}
\item \texttt{claude-opus-4.8} (Claude Opus 4.8), Anthropic.
\item \texttt{gpt-5.6} (GPT-5.6), OpenAI.
\item \texttt{grok-4.5} (Grok 4.5), xAI.
\item \texttt{deepseek-v4} (DeepSeek-V4), DeepSeek.
\item \texttt{kimi-k3} (Kimi K3), Moonshot AI.
\item \texttt{mistral-medium-3.5} (Mistral Medium 3.5), Mistral AI.
\end{itemize}
All outputs were collected directly from each provider's official API.
We requested temperature 0 and did not request any thinking or reasoning mode. When an endpoint rejected the temperature parameter, the request was sent again without it, so that model ran at its provider's default sampling temperature.
Every model received the same system prompt, bias definitions and output format in every setting (Appendix~\ref{app:prompts}), so differences between settings are attributable to the in-context examples alone. Even with temperature set to 0, repeated queries did not return identical outputs (Appendix~\ref{app:seeds}).

\section{Score Details}
\label{app:scores}
\label{app:exclusion}

\paragraph{From output to prediction.} Each model output is parsed as a JSON object with a document-level field \texttt{biases\_present} and a list of \texttt{triggers}, each holding a bias code and one or more span strings (Figure~\ref{fig:prompt-output}). Span strings are aligned to character offsets in the article as described in Appendix~\ref{app:alignment}. The prompt requires \texttt{biases\_present} to equal the set of biases among the triggers. When a model violates this rule, we use the union of the two sets as its Task~1 prediction. Outputs that cannot be parsed, or that parse but lack the required fields, are scored as empty predictions rather than dropped, so a model pays for every unusable output (Appendix~\ref{app:validity}).

\paragraph{Corpus-level scores.} Table~\ref{tab:main} reports corpus-level scores. For each model, setting and run, true positives, false positives and false negatives are pooled over all 50 articles and nine categories before computing $F_1$ (Equation~\ref{eq:f1-micro}). For $k>0$ we compute these scores separately for each of the three example draws and report their mean. Per-draw standard deviations are in Appendix~\ref{app:seeds}.

\paragraph{Per-article scores and tests.} Since the article is the sampling unit, all significance tests use per-article scores. For each model, setting and article we compute the same metrics on that article alone and average them over the three runs. An article with no gold trigger for which the model also predicts none receives a score of 1, since it is a correct answer that the $F_1$ convention would otherwise score 0. This affects only per-article scores, as such an article contributes no true positives, false positives or false negatives at the corpus level. Differences between conditions are tested with two-sided paired $t$-tests over the 50 articles, and the 95\% confidence intervals in Figures~\ref{fig:task1} and~\ref{fig:task2} are the corresponding paired $t$-intervals. We correct for multiple comparisons with the Holm--Bonferroni procedure, taking as family all bars of one figure (Figures~\ref{fig:task1} and~\ref{fig:task2}) or all pairs of models within one metric (Section~\ref{sec:res-bias}).

\paragraph{Thresholds.} We set $\tau_s = \tau_t = 0.5$ throughout. At this threshold, greedy one-to-one assignment is optimal for span-level matching, as no predicted span can have an IoU above 0.5 with two disjoint gold spans.

\section{Annotation Codebook}
\label{app:codebook}

The codebook below defines each of the nine cognitive bias triggers. Annotators worked from these entries. Each gives the category's definition, the cognitive mechanism it recruits, its textual trigger signals, a positive and a negative example, and notes on its boundary and on the category with which it is most readily confused. Examples are illustrative constructions used in annotator training.

\subsection*{B01 Illusory Truth}
\textbf{Definition.} A claim is repeated two or more times within the article, verbatim or in varied wording, such that repetition alone increases its perceived truth value independent of evidential support.
\textbf{Mechanism.} Processing fluency. Text encountered before is processed more easily, and that ease is misattributed as evidence of truth.
\textbf{Trigger signals.} Verbatim or near-verbatim repetition of a claim across paragraphs; reformulated repetition of the same proposition; anaphoric cues such as ``as noted earlier'' or ``to repeat''; echoing of the same claim across title, opening, and closing.
\textbf{Positive example.} ``The new tax will devastate small businesses. Local entrepreneurs warn that the policy will devastate small businesses across the country. As mentioned, this measure threatens to destroy small business in our communities.''
\textbf{Negative example.} ``The new tax will affect small businesses. According to a 2024 industry survey, 62\% of business owners expect reduced margins, and 18\% anticipate closures within two years.''
\textbf{Boundary and confusion.} Topical recurrence is not illusory truth. The trigger requires the same propositional content to be asserted repeatedly, not merely the same subject discussed from different angles. Most readily confused with B02. B01 concerns repetition of any claim regardless of valence, whereas B02 concerns asymmetric negative weighting.

\subsection*{B02 Negativity Bias}
\textbf{Definition.} Negative information (harm, threat, loss, failure) is foregrounded, elaborated, or weighted more heavily than positive or neutral information of comparable magnitude, biasing the reader's overall evaluation.
\textbf{Mechanism.} Negative valence captures attention disproportionately and is encoded with greater perceptual and semantic detail than positive valence of equal objective magnitude.
\textbf{Trigger signals.} Headline or lead foregrounding harm, loss, or threat; asymmetric elaboration of negative outcomes with comparable positives omitted or minimized; catastrophizing vocabulary (``crisis'', ``disaster'', ``collapse'', ``devastating''); loss-framed numerical reporting where gain framing was equally available.
\textbf{Positive example.} ``The economy shed 200,000 jobs last month in what experts called a catastrophic collapse. Meanwhile, a modest 30,000 new positions were added in tech.''
\textbf{Negative example.} ``The labor market lost 200,000 jobs in declining sectors and added 30,000 in expanding ones, a net contraction of 170,000 positions.''
\textbf{Boundary and confusion.} Reporting on a genuinely negative event is not in itself a trigger. The trigger requires asymmetric elaboration or catastrophizing relative to what the facts support. Distinguished from B01 by asymmetric weighting rather than repetition.

\subsection*{B03 Bandwagon Effect}
\textbf{Definition.} Claims are presented as believed or endorsed by a large or growing majority, triggering the tendency to align one's own belief with perceived social consensus.
\textbf{Mechanism.} Social proof. Majority adoption is treated as evidence of correctness, independent of the underlying argument.
\textbf{Trigger signals.} Explicit majority appeal (``most people'', ``everyone knows'', ``millions have already\ldots''); viral or trending language; polling data framed to emphasize consensus without methodological qualification; late-adoption shame framing (``you haven't heard about this yet?'').
\textbf{Positive example.} ``Millions of parents have already switched to the new formula. Are you still using the old one?''
\textbf{Negative example.} ``A survey of 1,000 parents found that 60\% preferred the new formula, though the authors note the sample was not nationally representative.''
\textbf{Boundary and confusion.} Citing polling data is not itself a trigger. The data must be framed to create social pressure or imply that deviating from the majority is unreasonable. Most readily confused with B04. B03 invokes the crowd and B04 a specific authority, and the two may co-occur in a single span.

\subsection*{B04 Authority Bias}
\textbf{Definition.} A claim is endorsed by or attributed to an authority figure or institution in a way that substitutes for, rather than accompanies, evidential support.
\textbf{Mechanism.} Source expertise, real or claimed, is used as a peripheral cue to accept the message without scrutinizing its content.
\textbf{Trigger signals.} Expert endorsement without cited evidence; credential inflation (``leading scientist'', ``world-renowned expert''); institutional authority invoked to close debate (``the science is settled''); irrelevant or off-domain authority presented with impressive credentials.
\textbf{Positive example.} ``Dr Helena Marsh, a leading nutrition expert, warns that the chemical changes in the new formula are `potentially devastating' for children.''
\textbf{Negative example.} ``Dr Helena Marsh, who led the 2023 longitudinal study published in The Lancet, reports that the modified compound was associated with a 12\% increase in adverse outcomes ($n = 4{,}200$, $p < 0.01$).''
\textbf{Boundary and confusion.} Quoting an expert is not itself a trigger. The credential must substitute for the evidential basis the claim would otherwise need. Distinguished from B03 by invoking a specific authority rather than the crowd.

\subsection*{B05 Stereotyping}
\textbf{Definition.} A group is described as having fixed traits, dispositions, or characteristics, with individual members presumed to share them by virtue of group membership.
\textbf{Mechanism.} Prototype matching. A group label activates a fixed set of attributed characteristics, and individual variation is suppressed.
\textbf{Trigger signals.} Generic group statements (``Women are\ldots'', ``Immigrants tend to\ldots''); essentialist framing (``it's in their nature'', ``cultural DNA''); statistical generalization applied at the individual level without hedging; traits attributed to group membership rather than to circumstance or structure.
\textbf{Positive example.} ``These migrants don't value education --- it's a cultural thing. That's why their children underperform.''
\textbf{Negative example.} ``Children from low-income households, regardless of national or cultural background, face documented systemic barriers in education, including under-resourced schools and reduced access to early-years support.''
\textbf{Boundary and confusion.} Properly hedged statistical claims (``on average'', ``in this sample'') are not stereotyping. The trigger is unqualified generalization or attribution of outcomes to group essence rather than to structural conditions.

\subsection*{B06 Declinism}
\textbf{Definition.} The past is systematically presented as superior to the present, framing current conditions as degraded, lost, or corrupted relative to an idealized prior state.
\textbf{Mechanism.} The idealized past serves as an anchor, and the present is evaluated against a selectively reconstructed, rosier baseline.
\textbf{Trigger signals.} Explicit nostalgia framing (``things were better when\ldots''); asymmetric comparison in which past virtues are detailed and past problems ignored; ``golden age'' language; restoration rhetoric (``return to'', ``reclaim what was lost'').
\textbf{Positive example.} ``There was a time when this country worked. People had real jobs, real communities, real values. Today we have nothing of the kind --- only decay where there was once dignity.''
\textbf{Negative example.} ``Manufacturing employment in the region fell from 38\% in 1970 to 11\% in 2023, while service-sector employment rose from 41\% to 67\% over the same period.''
\textbf{Boundary and confusion.} Acknowledging that something has worsened is not itself declinism. The trigger requires the present to be framed as degraded against an idealized past whose own problems are selectively omitted. Most readily confused with B09. Declinism is an evaluative comparison (the past was better), whereas hindsight is epistemic certainty (the outcome was predictable).

\subsection*{B07 Identifiable Victim Effect}
\textbf{Definition.} A single, named, humanized victim produces a far stronger emotional response and call to action than statistical descriptions of equal or greater harm affecting anonymous populations.
\textbf{Mechanism.} Concrete, specific individuals activate empathy more powerfully than statistical abstraction, so the identified life is weighted disproportionately to the statistical one.
\textbf{Trigger signals.} A named victim with biographical detail (name, age, family, backstory); first-person or direct-speech testimony of suffering; explicit contrast between the individual and a statistic (``this is not a number, this is Sarah''); emotional focus on one person while aggregate data is absent or minimized.
\textbf{Positive example.} ``Seven-year-old Emma loved football and drawing. Last Tuesday, she died from a preventable disease. How many more Emmas must we lose?''
\textbf{Negative example.} ``The disease kills approximately 4,000 children annually in this region. Emma's case, reported last week, reflects a pattern documented across 23 countries.''
\textbf{Boundary and confusion.} A personal story presented alongside statistical context does not trigger the bias. The trigger is disproportionate or exclusive use of the individual case to motivate a claim, with aggregate evidence suppressed or absent.

\subsection*{B08 Illusory Correlation}
\textbf{Definition.} A text presents two events or variables as causally or correlatively linked when the co-occurrence is coincidental, selective, or logically insufficient to establish the claimed relationship.
\textbf{Mechanism.} Pattern detection overgeneralizes. Distinctive co-occurrences are weighted more heavily than their joint frequency warrants, and temporal sequence is mistaken for causal mechanism.
\textbf{Trigger signals.} Post hoc reasoning (``after X, Y occurred, which is why\ldots''); juxtaposition of events with no stated mechanism; selective reporting of only the co-occurring cases; correlation presented as causation without mechanism, controls, or counterfactual.
\textbf{Positive example.} ``The day after the new mayor took office, crime rose by 5\%. His policies are already making the city less safe.''
\textbf{Negative example.} ``Crime statistics for the month following the new mayor's inauguration showed a 5\% increase. Criminologists note that month-to-month fluctuations of this magnitude are typical and cannot be attributed to policies implemented only days earlier.''
\textbf{Boundary and confusion.} Not every temporal sequence is a trigger. The text must present the sequence as evidence of a causal or systematic relationship without sufficient justification. Most readily confused with B09. B08 is a forward causal claim (cause $\rightarrow$ effect), B09 a backward claim of prior predictability (outcome $\rightarrow$ ``should have known'').

\subsection*{B09 Hindsight Bias}
\textbf{Definition.} Knowing the outcome of an event, a text presents that outcome as having been obvious or predictable beforehand, understating the genuine uncertainty that existed at the time.
\textbf{Mechanism.} Outcome knowledge is projected backwards onto the prior state, inflating perceived ex-ante predictability and erasing genuine epistemic uncertainty.
\textbf{Trigger signals.} Phrases asserting prior obviousness (``it was always clear that\ldots'', ``anyone could have seen this coming'', ``it was inevitable''); retrospective certainty reconstructing events as foreordained; blame framing dependent on post-hoc predictability (``they should have known''); selective citation of early warnings while ignoring contemporary uncertainty.
\textbf{Positive example.} ``It was obvious from the beginning that the policy would fail. The warning signs were all there, and anyone paying attention could have predicted this outcome.''
\textbf{Negative example.} ``Analysts were divided at the time of the policy's launch. Some predicted success, citing X, and others warned of failure, citing Y. In retrospect, the critics' concerns proved well-founded.''
\textbf{Boundary and confusion.} Legitimate retrospective analysis that acknowledges prior uncertainty is not hindsight bias. The trigger requires the outcome to be presented as having been obviously predictable, erasing the genuine ex-ante epistemic state. Distinguished from B06 by concerning certainty about a specific outcome rather than idealization of an era.

\section{Annotation Details}
\label{app:annotation}

\paragraph{Codebook freeze.} The codebook was frozen at the end of the two calibration rounds (Section~\ref{sec:annotators}), and neither the codebook nor the category set changed during annotation or adjudication.

\paragraph{Trigger grouping.} Spans of one discontinuous trigger share a group identifier. When the gold standard is loaded, spans connected by an annotated relation or carrying the same group tag are merged into one trigger, and grouping only ever applies within one bias. A passage labeled with two biases therefore contributes one span to a trigger of each bias.

\paragraph{Gold-standard statistics.} Table~\ref{tab:gold-stats} reports the gold standard at the level of triggers. Articles are on average 753 words long (median 463). Of the 398 gold triggers, 143 consist of more than one span, and 142 of these are Illusory Truth triggers. Outside B01, only one trigger (an Illusory Correlation trigger) spans more than one passage.

\begin{table*}[h]
\centering
\small
\setlength{\tabcolsep}{4pt}
\begin{tabular}{lrrrrr}
\toprule
Bias & Articles & Triggers & Multi-span & Spans & Avg.\ words \\
\midrule
B01 Illusory Truth             & 45 & 147 & 142 & 338 & 14.6 \\
B02 Negativity Bias            & 23 &  86 &   0 &  86 & 10.3 \\
B03 Bandwagon Effect           &  7 &  11 &   0 &  11 & 10.3 \\
B04 Authority Bias             & 19 &  34 &   0 &  34 & 28.1 \\
B05 Stereotyping               & 14 &  25 &   0 &  25 & 17.9 \\
B06 Declinism                  &  6 &  17 &   0 &  17 & 35.0 \\
B07 Identifiable Victim Effect &  7 &  20 &   0 &  20 & 26.4 \\
B08 Illusory Correlation       & 16 &  39 &   1 &  42 & 30.0 \\
B09 Hindsight Bias             & 11 &  19 &   0 &  19 & 13.4 \\
\midrule
Total                          & 46 & 398 & 143 & 592 & 16.9 \\
\bottomrule
\end{tabular}
\caption{Gold standard by bias. Articles: articles containing the bias (the total counts articles with at least one bias). Multi-span: triggers made of two or more spans. Avg.\ words: mean span length in words.}
\label{tab:gold-stats}
\end{table*}

\section{Prompts and In-Context Examples}
\label{app:prompts}

All settings share one system prompt (Figure~\ref{fig:prompt-system}). It describes the two tasks, gives a one-line definition of each bias, states the rule for multi-span triggers, and fixes the output format. The user message (Figure~\ref{fig:prompt-user}) contains the target article. With $k>0$, the target article is preceded by $k$ annotated example articles, which are part of the user message rather than separate dialogue turns. The examples are drawn separately for each target article from the other 49 articles, so that no article ever appears among its own examples, with a random generator seeded by the run seed (17, 23 or 29) and the target article. The draw is not fully uniform. One of the $k$ examples is always drawn from the articles that contain a multi-span trigger, so that every few-shot prompt demonstrates the grouping convention, and the remaining $k-1$ are drawn uniformly from the rest. The order of the examples is then shuffled. Each example consists of the full article text followed by its gold annotation in the required output format. In the examples, the \texttt{rationale} field is filled with a fixed sentence (``These span(s) jointly trigger B01 Illusory Truth.''), so the examples show which spans form a trigger but not why. An example article without triggers is shown with \texttt{biases\_present} set to \texttt{["NONE"]} and an empty trigger list.

\paragraph{Definitions seen by the models.} The models received the one-line definitions in Figure~\ref{fig:prompt-system}, not the annotator codebook of Appendix~\ref{app:codebook}. Some of the boundary conditions in the codebook appear in these definitions in compressed form, for instance ``disproportionate weight'' for Negativity Bias and ``beyond what the evidence supports'' for Authority Bias. Others do not. The definition of Declinism, for instance, does not say that the idealized past must omit its own problems. The prompt also asks that each span belong to exactly one trigger, whereas the gold standard lets one passage carry triggers of two biases (Appendix~\ref{app:annotation}).

\begin{figure*}[t]
\begin{promptbox}{System prompt}
\textbf{System message:}\\[2pt]
You are an expert annotator of cognitive bias triggers in news articles. Your task has two linked parts:
\begin{enumerate}\setlength{\itemsep}{0pt}
\item \textbf{Document-level detection}: identify which cognitive biases from the taxonomy below are triggered anywhere in the article.
\item \textbf{Trigger extraction}: for each bias you detect, extract the \emph{triggers} --- the specific text that would activate that bias in a reader. A trigger is a unit of persuasive effect, and it may consist of ONE span or SEVERAL related spans that jointly produce the effect.
\end{enumerate}
\textbf{Bias taxonomy}
\begin{itemize}\setlength{\itemsep}{0pt}
\item \textbf{B01 Illusory Truth} --- a claim gains credibility through repetition. The trigger is the \emph{set of repetitions}: the same claim or near-paraphrase stated at multiple points in the article.
\item \textbf{B02 Negativity Bias} --- disproportionate weight given to negative framing, threats, losses, or alarming language.
\item \textbf{B03 Bandwagon Effect} --- appeal to popularity or consensus (``everyone knows'', ``growing numbers agree'') as grounds for belief.
\item \textbf{B04 Authority Bias} --- reliance on an authority figure, title, or institution as the warrant for a claim, beyond what the evidence supports.
\item \textbf{B05 Stereotyping} --- attributing traits to individuals based on group membership.
\item \textbf{B06 Declinism} --- framing the present or future as decay from a better past.
\item \textbf{B07 Identifiable Victim Effect} --- a single named or vividly described individual stands in for a broader issue, driving emotional response.
\item \textbf{B08 Illusory Correlation} --- implying a causal or meaningful link between co-occurring events or traits without support.
\item \textbf{B09 Hindsight Bias} --- presenting past events as having been predictable or inevitable (``it was obvious this would happen'').
\end{itemize}
\textbf{Critical rule: multi-span triggers}\\
A single trigger can span multiple, non-adjacent passages. This is \emph{typical for B01 Illusory Truth}: if a claim is repeated three times across the article, that is \textbf{ONE trigger with three spans}, not three separate triggers. Group all repetitions of the same claim into one trigger. Other biases usually have single-span triggers, but group spans whenever they only work jointly.\\[3pt]
\textbf{Output format}\\
Return ONLY a JSON object, no markdown fences, no commentary:\\ \textit{(JSON format shown in Figure~\ref{fig:prompt-output})}\\[3pt]
Rules:
\begin{itemize}\setlength{\itemsep}{0pt}
\item Every span MUST be a verbatim, character-exact substring of the article text. Do not paraphrase, normalize whitespace, or fix typos.
\item Keep spans minimal: the shortest stretch of text that carries the trigger, typically a clause or sentence, not a whole paragraph.
\item \texttt{biases\_present} must exactly equal the set of biases appearing in \texttt{triggers}. If no bias is triggered, return \texttt{\{"biases\_present": ["NONE"], "triggers": []\}}.
\item Number triggers T1, T2, \ldots{} in order of first appearance in the article.
\item A span belongs to exactly one trigger.
\end{itemize}
\end{promptbox}
\caption{System prompt used in every setting, verbatim. The JSON format it contains is shown in Figure~\ref{fig:prompt-output}.}
\label{fig:prompt-system}
\end{figure*}

\begin{figure}[t]
\begin{promptbox}{User message}
\textbf{User message:}\\[2pt]
\textit{(only for $k>0$)}\\
\texttt{\#\# Few-shot examples}\\[2pt]
\textit{(repeated for each of the $k$ examples)}\\
\texttt{\#\#\# Example [i]}\\
\texttt{ARTICLE:}\\
\texttt{[EXAMPLE ARTICLE TEXT]}\\
\texttt{OUTPUT:}\\
\texttt{[GOLD ANNOTATION AS JSON]}\\[4pt]
\texttt{\#\# Target}\\
\texttt{ARTICLE:}\\
\texttt{[ARTICLE TEXT]}\\
\texttt{OUTPUT:}
\end{promptbox}
\caption{User message. The example block appears only in the few-shot settings.}
\label{fig:prompt-user}
\end{figure}

\begin{figure}[t]
\begin{promptbox}{Output format}
\small
\begin{verbatim}
{
  "biases_present": ["B01", "B02"],
  "triggers": [
    {
      "trigger_id": "T1",
      "bias": "B01",
      "spans": [
        "first verbatim occurrence of
         the repeated claim",
        "second verbatim occurrence,
         possibly reworded",
        "third verbatim occurrence"
      ],
      "rationale": "one sentence: why
        these spans jointly trigger
        this bias"
    },
    {
      "trigger_id": "T2",
      "bias": "B02",
      "spans": ["single verbatim span"],
      "rationale": "..."
    }
  ]
}
\end{verbatim}
\end{promptbox}
\caption{Output format as given in the system prompt (line breaks inside strings added for display). The spans of one trigger are listed together, which is how models express grouping for Task~2.}
\label{fig:prompt-output}
\end{figure}

\section{Output Validity}
\label{app:validity}

Table~\ref{tab:validity} counts the outputs that could not be used as returned, pooled over the three runs per setting (150 outputs per cell). Unusable outputs are rare except for GPT-5.6, for which 10\% of zero-shot and 13\% of $k=10$ outputs are unusable. Its failures concentrate on the same articles across runs. GPT-5.6 never returned a usable zero-shot output for article 48, and its outputs for articles 13, 30 and 40 lacked the required fields in all three zero-shot and all three $k=10$ runs. Because unusable outputs are scored as empty predictions, they lower GPT-5.6's recall in Table~\ref{tab:main}. The most frequent inconsistency, a \texttt{biases\_present} field that omits biases the model itself annotated, is most common for Claude Opus~4.8 with examples (12 of 150 outputs) and is resolved by taking the union of the two sets (Appendix~\ref{app:scores}).

\begin{table*}[h]
\centering
\small
\setlength{\tabcolsep}{4pt}
\begin{tabular}{lrrrrrr}
\toprule
 & \multicolumn{3}{c}{$k=0$} & \multicolumn{3}{c}{$k=10$} \\
\cmidrule(lr){2-4}\cmidrule(lr){5-7}
Model & No JSON & No fields & Mismatch & No JSON & No fields & Mismatch \\
\midrule
Claude Opus 4.8    & 0 &  0 & 1 & 0 &  0 & 12 \\
GPT-5.6            & 3 & 12 & 2 & 3 & 16 &  8 \\
Grok 4.5           & 0 &  1 & 0 & 0 &  0 &  3 \\
DeepSeek-V4        & 2 &  0 & 1 & 0 &  0 &  2 \\
Kimi K3            & 5 &  0 & 1 & 1 &  0 &  0 \\
Mistral Medium 3.5 & 0 &  0 & 0 & 2 &  6 &  3 \\
\bottomrule
\end{tabular}
\caption{Unusable and inconsistent outputs out of 150 per cell (three runs of 50 articles). No JSON: no valid JSON object could be parsed. No fields: a JSON object was recovered from surrounding text but lacked \texttt{biases\_present} and \texttt{triggers}. Both are scored as empty predictions. Mismatch: \texttt{biases\_present} differs from the biases among the triggers. The $k=3$ setting is omitted.}
\label{tab:validity}
\end{table*}

\section{Span Alignment}
\label{app:alignment}

Models return each span as a string copied from the article rather than as character offsets. Before scoring, we align every predicted string to the source text with a cascade of four matchers, stopping at the first that succeeds.
\begin{enumerate}
\item \textbf{Exact.} The string occurs verbatim in the article, as returned or after stripping leading and trailing whitespace.
\item \textbf{Case-normalized.} The string occurs in the article after lower-casing both. This recovers spans where the model changed the capitalization of the first word.
\item \textbf{Whitespace-flexible.} The words of the string occur in the article in the same order, separated by any whitespace (first case-sensitive, then case-insensitive). This recovers spans where the model replaced a line break in the article with a space.
\item \textbf{Fuzzy.} We find the longest block of characters that the string shares with the article, which must cover at least 10 characters and at least half the string. We then take the window of the article that has the string's length and lines up with that block, and accept it if its \texttt{difflib} similarity ratio with the string is at least 0.80.
\end{enumerate}
A string that none of the matchers can place is marked as unlocated. Unlocated spans are never discarded. At the character level they contribute their full length as false positives, at the span level they count as false positive spans, and at the trigger level their length is added to the size of the predicted trigger (Section~\ref{sec:tasks}).

\paragraph{Repeated strings.} A string can match the article in several places, which is common for Illusory Truth, where an identical claim recurs. The candidate positions are then ranked by three criteria in turn. A position not yet taken by an earlier span of the same bias in the same output comes first, so that repeated mentions are spread over their occurrences. Next comes a position that overlaps a gold span of the same bias, and last the leftmost position. The second criterion uses the gold standard, but only to choose between positions where the article contains exactly the string the model returned. It cannot turn a wrong string into a correct one, and without it the score would reflect the aligner's arbitrary choice of occurrence rather than the model's output. Copying the gold spans as predictions scores a character-level $F_1$ of 1.0.

\begin{table*}[h]
\centering
\small
\setlength{\tabcolsep}{4pt}
\begin{tabular}{lrrrrrr}
\toprule
Model & Spans & Exact & Case & Whitesp. & Fuzzy & Unloc. \\
\midrule
Claude Opus 4.8    & 2{,}590 & 99.19 & 0.31 & 0.00 & 0.46 & 0.04 \\
GPT-5.6            & 3{,}015 & 99.70 & 0.17 & 0.00 & 0.10 & 0.03 \\
Grok 4.5           & 4{,}288 & 99.00 & 0.23 & 0.05 & 0.65 & 0.07 \\
DeepSeek-V4        & 3{,}037 & 96.61 & 0.76 & 0.36 & 1.91 & 0.36 \\
Kimi K3            & 2{,}535 & 98.58 & 0.04 & 0.12 & 1.14 & 0.12 \\
Mistral Medium 3.5 & 4{,}004 & 95.05 & 1.62 & 0.85 & 2.07 & 0.40 \\
\midrule
Zero-shot          & 8{,}420 & 97.68 & 0.59 & 0.43 & 1.14 & 0.15 \\
$k=10$             & 11{,}049 & 98.05 & 0.56 & 0.13 & 1.06 & 0.20 \\
\bottomrule
\end{tabular}
\caption{Share of predicted spans (\%) resolved by each alignment step, pooled over the three zero-shot runs and the three $k=10$ example draws.}
\label{tab:alignment}
\end{table*}

Alignment is close to lossless. Every model copies at least 95\% of its spans verbatim, and at most 0.40\% of any model's spans, or 0.29\% of its predicted characters, cannot be located. DeepSeek-V4 and Mistral Medium~3.5 rely most on the later matching steps, which resolve 3.0\% and 4.5\% of their spans respectively.

Fuzzy matches mostly repair small copying errors, such as a changed verb tense or a dropped article. The median similarity between the model's string and the aligned text is 0.98. A small number of fuzzy matches attach a string that does not occur in the article to a similar passage, for instance a quotation in which the model altered a time expression but kept the rest of the sentence. Of the 213 fuzzy matches, 21 have a similarity below 0.90 to the text they are aligned to. These make up about 0.1\% of all predicted spans.

\section{Robustness to Example Draws and Reruns}
\label{app:seeds}

Table~\ref{tab:seed-sd} gives the standard deviation of each corpus-level score over the three runs per setting. For $k=3$ and $k=10$ this is the variation over the three example draws. Zero-shot prompts contain no examples, but each of the three runs re-queried the models, so the zero-shot standard deviation measures run-to-run nondeterminism. This nondeterminism is substantial at the level of individual outputs. Across the three zero-shot runs, a model returned exactly the same triggers for between 1 (Claude Opus~4.8, Grok~4.5) and 9 (GPT-5.6) of the 50 articles, and exactly the same set of biases for between 6 (Kimi~K3) and 35 (DeepSeek-V4). At the corpus level it largely averages out. All standard deviations are at most 0.042, and at most 0.024 for the Task~2 scores, which is small compared with the gains from ten examples for every model except Mistral Medium~3.5, whose gains are not significant (Figures~\ref{fig:task1} and~\ref{fig:task2}). Because of this nondeterminism, Table~\ref{tab:main} and Figures~\ref{fig:task1} and~\ref{fig:task2} average the zero-shot condition over all six zero-shot runs, one accompanying each few-shot run. The standard deviations in Table~\ref{tab:seed-sd} are computed over the three zero-shot runs that accompany the $k=10$ draws.

\begin{table*}[h]
\centering
\small
\setlength{\tabcolsep}{5pt}
\begin{tabular}{llccccc}
\toprule
 & & \multicolumn{2}{c}{Task 1} & \multicolumn{3}{c}{Task 2} \\
\cmidrule(lr){3-4}\cmidrule(lr){5-7}
Model & $k$ & Micro-F1 & Subset Acc. & Char-F1 & Span-F1 & Trigger-F1 \\
\midrule
\multirow{3}{*}{Claude Opus 4.8}    & 0  & 0.004 & 0.000 & 0.019 & 0.013 & 0.017 \\
                                    & 3  & 0.025 & 0.020 & 0.005 & 0.003 & 0.007 \\
                                    & 10 & 0.022 & 0.020 & 0.011 & 0.007 & 0.010 \\
\multirow{3}{*}{GPT-5.6}            & 0  & 0.006 & 0.012 & 0.015 & 0.003 & 0.016 \\
                                    & 3  & 0.036 & 0.042 & 0.024 & 0.017 & 0.021 \\
                                    & 10 & 0.021 & 0.031 & 0.019 & 0.016 & 0.012 \\
\multirow{3}{*}{Grok 4.5}           & 0  & 0.021 & 0.012 & 0.010 & 0.003 & 0.008 \\
                                    & 3  & 0.014 & 0.012 & 0.012 & 0.010 & 0.001 \\
                                    & 10 & 0.028 & 0.031 & 0.014 & 0.003 & 0.008 \\
\multirow{3}{*}{DeepSeek-V4}        & 0  & 0.013 & 0.000 & 0.008 & 0.003 & 0.004 \\
                                    & 3  & 0.010 & 0.012 & 0.015 & 0.008 & 0.011 \\
                                    & 10 & 0.010 & 0.023 & 0.010 & 0.009 & 0.006 \\
\multirow{3}{*}{Kimi K3}            & 0  & 0.011 & 0.023 & 0.008 & 0.007 & 0.012 \\
                                    & 3  & 0.006 & 0.012 & 0.010 & 0.009 & 0.009 \\
                                    & 10 & 0.029 & 0.012 & 0.010 & 0.002 & 0.006 \\
\multirow{3}{*}{Mistral Medium 3.5} & 0  & 0.005 & 0.000 & 0.015 & 0.009 & 0.009 \\
                                    & 3  & 0.018 & 0.020 & 0.020 & 0.015 & 0.012 \\
                                    & 10 & 0.027 & 0.023 & 0.007 & 0.006 & 0.009 \\
\bottomrule
\end{tabular}
\caption{Standard deviation of corpus-level scores over three runs. $k=0$: three reruns of the same prompt. $k=3$ and $k=10$: three example draws (seeds 17, 23, 29).}
\label{tab:seed-sd}
\end{table*}

\section{How Models Group Spans}
\label{app:grouping}

Trigger-F1 is the only metric that evaluates grouping (Section~\ref{sec:tasks}). Table~\ref{tab:grouping} describes how the models group spans into triggers. In the gold standard, grouping is almost entirely a property of Illusory Truth. 97\% of B01 triggers consist of several mentions of one claim, whereas only 1 of the 251 triggers of the other eight biases spans more than one passage.

Zero-shot, the models over-group. All of them group B01 mentions, as the definition asks, but they also merge passages of other biases into multi-span triggers, from 25\% of non-B01 triggers for Mistral Medium~3.5 to 80\% for DeepSeek-V4. They return fewer triggers than the gold standard (3.1 to 6.1 per article against 8.0) and pack more mentions into each B01 trigger (2.8 to 3.7 against 2.3). Examples largely correct this. At $k=10$, the share of multi-span triggers outside B01 falls to between 1\% and 11\% for five of the six models, and the number of triggers per article rises to between 6.0 and 11.3, around the gold value. Claude Opus~4.8 changes least and still groups 23\% of its non-B01 triggers. This change in grouping coincides with the trigger-F1 gains from examples (Figure~\ref{fig:task2}), which suggests that the examples convey BAIT's grouping convention. Span length, by contrast, already matches the gold standard zero-shot. Predicted spans average 14 to 18 words against 17 for gold spans, in both settings.

\begin{table*}[h]
\centering
\small
\setlength{\tabcolsep}{4pt}
\begin{tabular}{llrrrrr}
\toprule
 & & Triggers & Spans & Spans per & \multicolumn{2}{c}{Multi-span (\%)} \\
\cmidrule(lr){6-7}
Model & $k$ & per article & per trigger & B01 trigger & B01 & Other \\
\midrule
Gold standard & -- & 7.96 & 1.49 & 2.30 & 96.6 & 0.4 \\
\midrule
\multirow{2}{*}{Claude Opus 4.8}    & 0  & 3.47 & 2.02 & 3.34 & 100.0 & 58.5 \\
                                    & 10 & 5.97 & 1.72 & 2.85 & 100.0 & 23.3 \\
\multirow{2}{*}{GPT-5.6}            & 0  & 4.94 & 1.71 & 3.30 & 100.0 & 26.5 \\
                                    & 10 & 7.20 & 1.62 & 2.97 & 100.0 &  6.3 \\
\multirow{2}{*}{Grok 4.5}           & 0  & 5.39 & 2.47 & 3.72 & 100.0 & 55.0 \\
                                    & 10 & 10.02 & 1.52 & 2.67 & 97.5 &  4.1 \\
\multirow{2}{*}{DeepSeek-V4}        & 0  & 3.06 & 2.86 & 3.17 & 100.0 & 80.1 \\
                                    & 10 & 8.23 & 1.40 & 2.27 & 99.7 &  5.0 \\
\multirow{2}{*}{Kimi K3}            & 0  & 3.73 & 1.89 & 3.15 & 100.0 & 37.4 \\
                                    & 10 & 6.11 & 1.61 & 2.68 & 100.0 & 10.5 \\
\multirow{2}{*}{Mistral Medium 3.5} & 0  & 6.09 & 1.90 & 2.83 & 100.0 & 24.9 \\
                                    & 10 & 11.31 & 1.34 & 2.11 & 97.2 &  0.6 \\
\bottomrule
\end{tabular}
\caption{Grouping of predicted spans into triggers, averaged over the three runs per setting (150 outputs). Multi-span: share of triggers with two or more spans, for Illusory Truth (B01) and for the other eight biases. Unusable outputs (Appendix~\ref{app:validity}) count as articles without triggers.}
\label{tab:grouping}
\end{table*}

\end{document}